\documentclass[11pt]{lmcs}
\pdfoutput=1

\usepackage[utf8]{inputenc}

\usepackage{lastpage}
\lmcsdoi{20}{2}{2}
\lmcsheading{}{\pageref{LastPage}}{}{}%
{Mar.~10,~2023}{Apr.~09,~2024}{}

\def\isdraft{0}

\keywords{Analogical Proportions, Analogical Reasoning, Boolean Algebra, Universal Algebra, Logic in Computer Science}

\usepackage{graphicx}
\usepackage{bbm}
\usepackage{mathtools}
\usepackage{amsmath,amssymb,amsfonts,dsfont}
\usepackage{prettyref} 
\usepackage{enumitem}
\usepackage[hyphens]{url}
\usepackage{tikz-cd}
\usepackage{tikz}
\usetikzlibrary{positioning}
\usetikzlibrary{arrows}
\usepackage{hyperref} 
\usepackage[color=black,textcolor=white\if\isdraft0,disable\fi]{todonotes}
\usepackage{stackengine}
\usepackage[normalem]{ulem}
\usepackage{stmaryrd}
\usepackage{pdfpages}
\usepackage{marginnote}
\usepackage{float}
\graphicspath{{fig/}}
\usetikzlibrary{arrows,automata,topaths,matrix,positioning,fit}
\usepgflibrary{shapes.geometric}
\usetikzlibrary{shapes.geometric}
\tikzset{every state/.style={minimum size=0pt}}
\usepackage{footnote}
\usepackage{float}
\usepackage{tabularx}
\usepackage{booktabs}

\newrefformat{§}{Section \ref{#1}}
\newrefformat{appendix}{Appendix \ref{#1}}
\newrefformat{c}{Corollary \ref{#1}}
\newrefformat{equ}{(\ref{#1})}
\newrefformat{e}{Example \ref{#1}}
\newrefformat{exe}{Exercise \ref{#1}}
\newrefformat{d}{Definition \ref{#1}}
\newrefformat{f}{Fact \ref{#1}}
\newrefformat{fig}{Figure \ref{#1}}
\newrefformat{i}{Item \ref{#1}}
\newrefformat{l}{Lemma \ref{#1}}
\newrefformat{note}{Note \ref{#1}}
\newrefformat{n}{Notation \ref{#1}}
\newrefformat{o}{Observation \ref{#1}}
\newrefformat{problem}{Problem \ref{#1}}
\newrefformat{p}{Proposition \ref{#1}}
\newrefformat{pseudocode}{Pseudocode \ref{#1}}
\newrefformat{r}{Remark \ref{#1}}
\newrefformat{table}{Table \ref{#1}}
\newrefformat{t}{Theorem \ref{#1}}

\newcommand{\righttherefore}{:\joinrel\cdot\,}

\def\enforceminheight#1{\llap{\phantom{\rule{0pt}{#1}}}}

\begin{document}

\title{
	Boolean proportions
}
\thanks{
	I thank the anonymous reviewers for their valuable suggestions.
}
\author{
	Christian Anti\'c\lmcsorcid{0000-0001-9338-9489}
}
\address{
	Vienna University of Technology\\
	Vienna, Austria
}
\email{
	christian.antic@icloud.com
}

\begin{abstract}
	The author has recently introduced an abstract algebraic framework of analogical proportions within the general setting of universal algebra. This paper studies analogical proportions in the boolean domain consisting of two elements 0 and 1 within his framework. It turns out that our notion of boolean proportions coincides with two prominent models from the literature in different settings. This means that we can capture two separate modellings of boolean proportions within a single framework which is mathematically appealing and provides further evidence for the robustness and applicability of the general framework.
\end{abstract}

\maketitle

\section{Introduction}

Analogical reasoning is the ability to detect parallels between two seemingly distant objects or situations, a fundamental human capacity used for example in commonsense reasoning, learning, and creativity which is believed by many researchers to be at the core of human and artificial general intelligence \cite{Gust08,Hofstadter13,Krieger03,Polya54}. Analogical proportions are expressions of the form ``$a$ is to $b$ what $c$ is to $d$'' at the core of analogical reasoning with applications in artificial intelligence \cite{Prade21}. The author has recently introduced an abstract algebraic framework of analogical proportions in the general setting of universal algebra \cite{Antic22}. It is a promising novel model of analogical proportions with appealing mathematical properties. The \textbf{purpose of this paper} is to study that framework in the boolean domain consisting of the booleans 0 and 1 together with logical functions, where the \textbf{motivation} is to better understand the general framework by \textit{fully} understanding it in the simple but nonetheless important boolean domain. Boolean proportions as studied in this paper provide the basis for boolean \textit{function} proportions and finite set proportions to be studied in the future (see \prettyref{§:FW}).

The \textbf{main technical part} of the paper is divided into four sections §\ref{§:Constants}--\ref{§:PL} studying concrete boolean domains and a fifth \prettyref{§:Comparison} comparing our model to Klein's \cite{Klein82} and Miclet's \& Prade's \cite{Miclet09} prominent frameworks. Interestingly, it turns out that our model coincides with Klein's model in case negation is representable, and with Miclet's \& Prade's otherwise. 

In \prettyref{§:Constants}, we begin by studying boolean proportions in the algebras $(\{0,1\})$, $(\{0,1\},0)$, $(\{0,1\},1)$, and $(\{0,1\},0,1)$, consisting only of the underlying boolean universe together with a (possibly empty) list of boolean constants and no functions. We prove in \prettyref{t:B} the following simple characterization of boolean proportions in $(\{0,1\})$:
\begin{align*} 
	(\{0,1\})\models a:b::c:d \quad\Leftrightarrow\quad (a=b \quad\text{and}\quad c=d) \quad\text{or}\quad (a\neq b \quad\text{and}\quad c\neq d).
\end{align*} Already in this simple case without functions an interesting phenomenon emerges which has been already observed in \cite[Theorem 28]{Antic22}: it turns out that boolean proportions are \textit{non-monotonic} in the sense that, for example, $1:0::0:1$ holds in $(\{0,1\})$ and fails in the expanded structures $(\{0,1\},0)$, $(\{0,1\},1)$, and $(\{0,1\},0,1)$ containing constants. This may have interesting connections to non-monotonic reasoning, which itself is crucial for commonsense reasoning and which has been prominently formalized within the field of answer set programming \cite{Lifschitz19}.

In \prettyref{§:Negation}, we then study the structure $(\{0,1\},\neg,B)$ containing negation, where $B$ is any set of constants from $\{0,1\}$. Surprisingly, we can show in \prettyref{t:neg} that $(\{0,1\},\neg,B)$ is equivalent to $(\{0,1\})$ with respect to boolean proportions, that is, we derive
\begin{align*} (\{0,1\},\neg,B)\models a:b::c:d \quad\Leftrightarrow\quad (\{0,1\})\models a:b::c:d.
\end{align*} 

In \prettyref{§:XOR}, we study boolean proportions with respect to the exclusive or operation. Interestingly, it turns out that in case the underlying algebra contains the constant $1$, the equation $x+1=\neg x$ shows that we can represent the negation operation, which then yields
\begin{align*} 
	(\{0,1\},+,1)\models a:b::c:d \quad\Leftrightarrow\quad (\{0,1\},\neg)\models a:b::c:d.
\end{align*}

In \prettyref{§:PL}, we then study the structure $(\{0,1\},\lor,\neg,B)$, where $B$ is again any set of constants from $\{0,1\}$. Since disjunction and negation are sufficient to represent all boolean functions, these structures employ full propositional logic. Even more surprisingly than in the case of negation before, we show in \prettyref{t:lor_neg} that the structure $(\{0,1\},\lor,\neg,B)$ containing \textit{all} boolean functions is again equivalent to the boolean structure $(\{0,1\})$ with respect to boolean proportions, that is, we derive
\begin{align*} 
	(\{0,1\},\lor,\neg,B)\models a:b::c:d \quad\Leftrightarrow\quad (\{0,1\})\models a:b::c:d.
\end{align*} This has interesting consequences. For example, in \prettyref{§:Comparison} we show that, somewhat unexpectedly, our notion of boolean proportion coincides with Klein's characterization in most cases and with Miclet's \& Prade's in some cases. This is interesting as our model is an instance of an abstract algebraic model not geared towards the boolean domain.

In the tradition of the ancient Greeks, Lepage \cite{Lepage03} proposes four properties, namely\footnote{\cite{Lepage03} uses different names for the properties; we remain here consistent with the nomenclature used in \cite[§4.2]{Antic22}.}
\begin{enumerate}
	\item symmetry: $a:b::c:d \quad\Leftrightarrow\quad c:d::a:b$,
	\item central permutation: $a:b::c:d \quad\Leftrightarrow\quad a:c::b:d$,
	\item strong inner reflexivity: $a:a::c:d \quad\Rightarrow\quad d=c$,
	\item strong inner reflexivity: $a:b::a:d \quad\Rightarrow\quad d=b$
\end{enumerate} as a guideline for formal models of analogical proportions. In \cite[Theorem 28]{Antic22} the author has argued why all of Lepage's properties except for symmetry cannot be assumed in general. In \cite[§4.3]{Antic22}, the author adapts Lepage's list of properties by preserving his symmetry axiom from above, and by adding (some of the properties have been considered by other authors as well):
\begin{enumerate}
	\item inner symmetry: $a:b::c:d \quad\Leftrightarrow\quad b:a::d:c$,
	\item inner reflexivity: $a:a::c:c$,
	\item reflexivity: $a:b::a:b$,
	\item determinism: $a:a::a:d \quad\Leftrightarrow\quad a=d$,
\end{enumerate} and proving that these properties are always satisfied within his framework. Notice that inner reflexivity and reflexivity are weak forms of Lepage's strong inner reflexivity and strong reflexivity properties, respectively, whereas inner symmetry is an `inner' variant of Lepage's symmetry axiom which requires symmetry to hold within the respective structures. Moreover, he considers the following properties (some of which have been considered by other authors as well):
\begin{enumerate}
	\item commutativity: $a:b::b:a$,
	\item transitivity: $a:b::c:d \quad\text{and}\quad c:d::e:f \quad\Rightarrow\quad a:b::e:f$,
	\item inner transitivity: $a:b::c:d \quad\text{and}\quad b:e::d:f \quad\Rightarrow\quad a:e::c:f$,
	\item central transitivity: $a:b::b:c \quad\text{and}\quad b:c::c:d \quad\Rightarrow\quad a:b::c:d$.
\end{enumerate} In this paper we prove in \prettyref{t:lor_neg} that, contrary to the general case \cite[Theorem 28]{Antic22}, in propositional logic all of the aforementioned properties --- including Lepage's --- are satisfied.


\section{Preliminaries}\label{§:P}

We denote the \textit{\textbf{booleans}} by 0 and 1, and define
\begin{align*} 
	0\lor 0&:=0 \quad\text{and}\quad 1\lor 0:=0\lor 1:=1\lor 1:=1\quad\text{(disjunction)},\\
	\neg 0&:=1 \quad\text{and}\quad \neg 1:=0\quad\text{(negation)},\\
	0+0&:=1+1:=0 \quad\text{and}\quad 0+1:=1+0:=1 \quad\text{(exclusive or)},\\
	x\land y&:=\neg(\neg x\lor\neg y) \quad\text{(conjunction)},\\
	x\supset y&:=\neg x\lor y \quad\text{(implication)},\\
	x\equiv y&:=(x\supset y)\land (y\supset x) \quad\text{(equivalence)}.
\end{align*} 

In this paper, a \textit{\textbf{boolean domain}} is any structure $(\{0,1\},F,B)$, where $F$ is a (possibly empty) subset of $\{\lor,\neg\}$ called \textit{\textbf{boolean functions}}, and $B$ is a (possibly empty) subset of the booleans in $\{0,1\}$ called \textit{\textbf{constants}} or \textit{\textbf{distinguished elements}}. It is well-known that every boolean function can be expressed in terms of disjunction and negation, which means that in the structure $(\{0,1\},\lor,\neg)$ we can employ full propositional logic. We write $(\{0,1\},F,B)\subseteq (\{0,1\},F',B')$ if $F\subseteq F'$ and $B\subseteq B'$.

In the rest of the paper, $\mathfrak B$ denotes a generic 2-element boolean domain $(\{0,1\},F,B)$. Given a boolean domain $\mathfrak B=(\{0,1\},F,B)$, a \textit{\textbf{formula}} over $\mathfrak B$ (or \textit{\textbf{$\mathfrak B$-formula}}) is any well-formed expression containing boolean operations from $F$, constants from $B$, and variables from a denumerable set of \textit{\textbf{variables}} $X$. Logical equivalence between two formulas is defined as usual. 

We will write $s(\mathbf x)$ in case $s$ is a formula containing variables among $\mathbf x$.\footnote{We use here ``$s$'' to denote a (propositional) formula --- instead of the more common notation using Greek letters $\varphi,\psi,\ldots$ --- since formulas are \textit{terms} in our algebraic setting.} Given a sequence of booleans $\mathbf a$ of same length as $\mathbf x$, we denote by $s(\mathbf a)$ the formula obtained from $s(\mathbf x)$ by substituting $\mathbf a$ for $\mathbf x$ in the obvious way. We call a formula $s$ \textit{\textbf{satisfiable}} (resp., \textit{\textbf{falsifiable}}) if there is some sequence of booleans $\mathbf a$ such that $s(\mathbf a)$ is logically equivalent to $1$ (resp., to $0$). Every formula $s(x_1,\ldots,x_n)$ induces a boolean function $s^\mathfrak B:\mathfrak B^n\to\mathfrak B$ in the obvious way by replacing the variables $x_1,\ldots,x_n$ by concrete values $a_1,\ldots,a_n\in\{0,1\}$. We call $s$ \textit{\textbf{injective}} in $\mathfrak B$ if $s^\mathfrak B$ is an injective function. The \textit{\textbf{rank}} of a formula $s$ is given by the number of variables occurring in $s$ denoted by $r(s)$.

Notice that every formula not containing variables is logically equivalent to a boolean, and with a slight abuse of notation we will not distinguish between logically equivalent formulas: for example, $x$ and $\neg\neg x$ denote the same formula et cetera.

We now instantiate the abstract algebraic framework of analogical proportions in \cite{Antic22} in the boolean domain. We want to functionally relate booleans via term rewrite rules as follows. Transforming $0$ to $1$ means, for example, that $x$ gets transformed into $\neg x$ which we can state more pictorially as the term rewrite rule $x\to\neg x$. Now transforming the boolean $1$ ``in the same way'' means to transform $1$ into $\neg 1$ which again is an instance of $x\to\neg x$. Let us make this notation official:

\begin{defi} We will always write $s\to t$ instead of $(s,t)$, for any pair of $\mathfrak B$-formulas $s$ and $t$ such that every variable in $t$ occurs in $s$.\footnote{Since rules of the form $s\to t$ ought to formalize the intuition that $s$ gets transformed into $t$, we need the restriction on variables to guarantee that only parts of $s$ get transformed.} We call such expressions \textit{\textbf{$\mathfrak B$-rewrite rules}} or \textit{\textbf{$\mathfrak B$-justifications}} and we often omit the reference to $\mathfrak B$. We denote the set of all such $\mathfrak B$-justifications by $J_\mathfrak B(X)$.
\end{defi}

The following definition --- which is an instance of a more general definition given in the setting of universal algebra \cite[Definition 8]{Antic22} --- is motivated by the observation that analogical proportions of the form $a:b::c:d$ are best defined in terms of arrow proportions $a\to b\righttherefore c\to d$ formalizing \textit{directed} transformations and a maximality condition on the set of justifications. More precisely, to say that ``$a$ transforms into $b$ as $c$ transforms into $d$'' means that the set of justifications $s\to t$ such that $a\to b$ and $c\to d$ are instances of $s\to t$ is maximal with respect to $d$, which intuitively means that the transformation $a\to b$ is maximally similar to the transformation $c\to d$.

\begin{defi} We define boolean proportions as follows:
\begin{enumerate}
	\item Define the \textit{\textbf{set of justifications}} of an \textit{\textbf{arrow}} $a\to b$ in $\mathfrak B$ by\footnote{We use here ``$\uparrow$'' instead of the original ``$Jus$'' in \cite{Antic22} as introduced in \cite{Antic23-22}.}
	\begin{align*} 
		\uparrow_\mathfrak B(a\to b):=\left\{s\to t\in J_\mathfrak B(X) \;\middle|\; a= s(\mathbf o)\text{ and }b= t(\mathbf o),\text{ for some }\mathbf o\in\{0,1\}^{r(s)}\right\}.
	\end{align*}

	\item Define the \textit{\textbf{set of justifications}} of an \textit{\textbf{arrow proportion}} $a\to b\righttherefore c\to d$\footnote{Read as ``$a$ transforms into $b$ as $c$ transforms into $d$''.} in $\mathfrak B$ by
	\begin{align*} 
		\uparrow_\mathfrak B(a\to b\righttherefore c\to d):=\ \uparrow_\mathfrak B(a\to b)\ \cap \uparrow_\mathfrak B(c\to d).
	\end{align*} We say that $J$ is a \textit{\textbf{trivial set of justifications}} in $\mathfrak B$ iff every justification in $J$ justifies every arrow proportion $a\to b\righttherefore c\to d$ in $\mathfrak B$, that is, iff
	\begin{align*} 
		J\subseteq\ \uparrow_\mathfrak B(a\to b\righttherefore c\to d)\quad\text{for all $a,b,c,d\in\{0,1\}$.}
	\end{align*} In this case, we call every justification in $J$ a \textit{\textbf{trivial justification}} in $\mathfrak B$. Now we say that the arrow proportion $a\to b\righttherefore c\to d$ \textit{\textbf{holds}} in $\mathfrak B$ written
	\begin{align*} 
		\mathfrak B \models a\to b\righttherefore c\to d
	\end{align*} iff
	\begin{enumerate}
		\item either $\uparrow_\mathfrak B(a\to b)\ \cup \uparrow_\mathfrak B(c\to d)$ consists only of trivial justifications, in which case there is neither a non-trivial transformation of $a$ into $b$ nor of $c$ into $d$;
		\item or $\uparrow_\mathfrak B(a\to b\righttherefore c\to d)$ is maximal with respect to subset inclusion among the sets $\uparrow_\mathfrak B(a\to b\righttherefore c\to d')$, $d'\in\{0,1\}$, containing at least one non-trivial justification, that is, for any element $d'\in \{0,1\}$,
		\begin{align*} 
			\{\text{trivial jus.}\}\subsetneq\ \uparrow_\mathfrak B(a\to b\righttherefore c\to d)&\subseteq\ \uparrow_\mathfrak B(a\to b\righttherefore c\to d')
		\end{align*} implies
		\begin{align*} 
			\{\text{trivial jus.}\}\subsetneq\ \uparrow_\mathfrak B(a\to b\righttherefore c\to d')\subseteq\ \uparrow_\mathfrak B(a\to b\righttherefore c\to d).
		\end{align*}
	\end{enumerate}

	\item\label{i:abcd} Finally, the boolean proportion relation is most succinctly defined by
	\begin{align*} 
		\mathfrak B \models a:b::c:d \quad:\Leftrightarrow\quad 
            &\mathfrak B \models a\to b\righttherefore c\to d \quad\text{and}\quad \mathfrak B \models b\to a\righttherefore d\to c\\
            &\mathfrak B \models c\to d\righttherefore a\to b \quad\text{and}\quad \mathfrak B \models d\to c\righttherefore b\to a.
	\end{align*}
\end{enumerate}
\end{defi}


\begin{rem} We will always write sets of justifications modulo renaming of variables, e.g., we will write $\{x\to x\}$ instead of $\{x\to x\mid x\in X\}$.
\end{rem}

\begin{defi} We call a $\mathfrak B$-formula $s$ a \textit{\textbf{generalization}} of a boolean $a$ in $\mathfrak B$ iff $a= s(\mathbf o)$, for some $\mathbf o\in\{0,1\}^{r(s)}$, and we denote the set of all generalizations of $a$ in $\mathfrak B$ by $\uparrow_\mathfrak B a$. Moreover, we define for any booleans $a,c\in\{0,1\}$:
\begin{align*} 
	a\uparrow_\mathfrak B c:=(\uparrow_\mathfrak B a)\cap (\uparrow_\mathfrak B c).
\end{align*} In particular, we have
\begin{align*} 
	\uparrow_\mathfrak B 0&=\{ s\mid s\text{ is falsifiable}\},\\
	\uparrow_\mathfrak B 1&=\{ s\mid s\text{ is satisfiable}\},\\
	0\uparrow_\mathfrak B 1&=\{ s\mid s\text{ is falsifiable and satisfiable}\}.
\end{align*}
\end{defi}

\begin{rem}\label{r:abcd} Notice that any justification $s\to t$ of $a\to b\righttherefore c\to d$ in $\mathfrak B$ must satisfy
\begin{align}\label{equ:varphi_psi} 
	a=s(\mathbf o) \quad\text{and}\quad b=t(\mathbf o) \quad\text{and}\quad c=s(\mathbf u) \quad\text{and}\quad d= t(\mathbf u),
\end{align} for some $\mathbf o, \mathbf u\in\{0,1\}^{r(s)}$. In particular, this means
\begin{align*}  
	s\in a\uparrow_\mathfrak B c \quad\text{and}\quad  t\in b\uparrow_\mathfrak B d.
\end{align*} We sometimes write $s\xrightarrow{\mathbf o\to\mathbf u} t$ to make the \textit{\textbf{witnesses}} $\mathbf o,\mathbf u$ and their transition explicit. This situation can be depicted as follows:
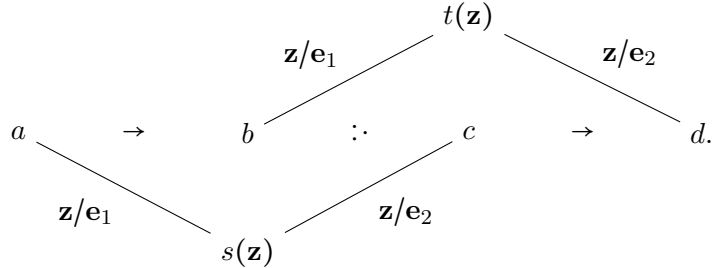
\begin{center}
\begin{tikzpicture}
	\node (a)               {$a$};
	\node (d1) [right=of a] {$\to $};
	\node (b) [right=of d1] {$b$};
	\node (d2) [right=of b] {$\righttherefore$};
	\node (c) [right=of d2] {$c$};
	\node (d3) [right=of c] {$\to $};
	\node (d) [right=of d3] {$d.$};
	\node (s) [below=of b] {$s(\mathbf x)$};
	\node (t) [above=of c] {$ t(\mathbf x)$};

	\draw (a) to [edge label'={$\mathbf x/\mathbf o$}] (s); 
	\draw (c) to [edge label={$\mathbf x/\mathbf u$}] (s);
	\draw (b) to [edge label={$\mathbf x/\mathbf o$}] (t);
	\draw (d) to [edge label'={$\mathbf x/\mathbf u$}] (t);
\end{tikzpicture}
\end{center} 

\end{rem}

The following reasoning pattern --- which roughly says that \textit{functional dependencies} are preserved under some conditions --- will often be used in the rest of the paper; it is a special case of \cite[Theorem 24]{Antic22}.

\begin{thm}[Functional Proportion Theorem]\label{t:FPT} In case $t$ is injective in $\mathfrak B$, we have
\begin{align*} 
	\mathfrak B\models a: t^\mathfrak B(a)::c: t^\mathfrak B(c).
\end{align*} In this case, we call $ t^\mathfrak B(c)$ a \textit{\textbf{functional solution}} of $a: t^\mathfrak B(a)::c:x$ in $\mathfrak B$ characteristically justified by $x\to t(x)$.
\end{thm}

Functional solutions are plausible since transforming $a$ into $ t(a)$ and $c$ into $ t(c)$ is a direct implementation of `transforming $a$ and $c$ in the same way'.


\cite{Lepage03} introduces the following postulates:
\begin{align}
	&\mathfrak B\models a:b::c:d\quad\Leftrightarrow\quad \mathfrak B\models c:d::a:b\quad\text{(symmetry)},\\
	\label{equ:central_permutation} &\mathfrak B\models a:b::c:d\quad\Leftrightarrow\quad \mathfrak B\models a:c::b:d\quad\text{(central permutation)},\\
	\label{equ:strong_inner_reflexivity} &\mathfrak B\models a:a::c:d\quad\Rightarrow\quad d=c\quad\text{(strong inner reflexivity)},\\
	\label{equ:strong_reflexivity} &\mathfrak B\models a:b::a:d\quad\Rightarrow\quad d=b\quad\text{(strong reflexivity)}.
\end{align} We will see in the forthcoming sections that all of Lepage's properties hold in the boolean setting studied here.

Although Lepage's properties appear reasonable in the boolean domain, \cite[Theorem 28]{Antic22} provides simple counter-examples to each of his properties (except for symmetry) in the \textit{general} case; he therefore considers the following alternative set of properties as a guideline for formal models of analogical proportions, adapted here to the boolean setting:
\begin{align}
	\label{equ:symmetry} \mathfrak B&\models a:b::c:d\quad\Leftrightarrow\quad\mathfrak B\models c:d::a:b\quad\text{(symmetry)},\\
	\label{equ:inner_symmetry} \mathfrak B&\models a:b::c:d \quad\Leftrightarrow\quad \mathfrak B\models b:a::d:c\quad\text{(inner symmetry)},\\
	\label{equ:inner_reflexivity} \mathfrak B&\models a:a::c:c\quad\text{(inner reflexivity)},\\
	\label{equ:reflexivity} \mathfrak B&\models a:b::a:b\quad\text{(reflexivity)},\\
	\label{equ:determinism} \mathfrak B&\models a:a::a:d \quad\Leftrightarrow\quad d=a\quad\text{(determinism)}.
\end{align} Moreover, the following properties are considered:
\begin{align} 
	&\label{equ:commutativity} \mathfrak B\models a:b::b:a\quad\text{(commutativity),}\\
	&\mathfrak B\models a:b::c:d \quad\text{and}\quad \mathfrak B\models c:d::e:f \quad\Rightarrow\quad \mathfrak B\models a:b::e:f \quad\text{(transitivity)},\\
	&\label{equ:inner_transitivity} \mathfrak B\models a:b::c:d \quad\text{and}\quad \mathfrak B\models b:e::d:f \quad\Rightarrow\quad \mathfrak B\models a:e::c:f \quad\text{(inner transitivity)},\\
	&\label{equ:central_transitivity} \mathfrak B\models a:b::b:c \quad\text{and}\quad \mathfrak B\models b:c::c:d \quad\Rightarrow\quad \mathfrak B\models a:b::c:d \quad\text{(central transitivity)},\\
	&\mathfrak B\models a:b::c:d \quad\text{and}\quad \mathfrak B\subseteq\mathfrak B' \quad\Rightarrow\quad \mathfrak B'\models a:b::c:d \quad\text{(monotonicity)}.
\end{align}

\begin{thm}\label{t:p} The boolean proportion relation \ref{i:abcd} satisfies symmetry \prettyref{equ:symmetry}, inner symmetry \prettyref{equ:inner_symmetry}, inner reflexivity \prettyref{equ:inner_reflexivity}, reflexivity \prettyref{equ:reflexivity}, and determinism \prettyref{equ:determinism}.
\end{thm}
\begin{proof} An instance of \cite[Theorem 28]{Antic22}.
\end{proof}

\section{Constants}\label{§:Constants}

In this section, we study boolean proportions in the structures $(\{0,1\})$, $(\{0,1\},0)$, $(\{0,1\},1)$, and $(\{0,1\},0,1)$, consisting only of the boolean universe $\{0,1\}$ and constants among 0 and 1 without boolean functions. This special case is interesting as it demonstrates subtle differences between structures containing different constants --- i.e., distinguished elements with a `name' in the language --- and here it makes a difference whether an element has a `name' or not.

Let us first say a few words about justifications in such structures. Recall that justifications are formula rewrite rules of the form $s\to t$. In the structure $(\{0,1\},B)$, consisting only of the booleans $0$ and $1$ with no functions and with the distinguished elements in $B\subseteq\{0,1\}$, each $(\{0,1\},B)$-formula is either a constant boolean from $B$ or a variable. The justifications in $(\{0,1\},B)$ can thus have only one of the following forms:
\begin{enumerate}
	\item The justification $x\to x$ justifies only directed variants of inner reflexivity \prettyref{equ:inner_reflexivity} of the form $a\to a\righttherefore c\to c$.
	\item The justification $x\to b$ justifies directed proportions of the form $a\to b\righttherefore c\to b$.
	\item The justification $a\to b$ justifies only directed variants of reflexivity \prettyref{equ:reflexivity} of the form $a\to b\righttherefore a\to b$.
\end{enumerate} The first case is the most interesting one as it shows that we can detect equality in $(\{0,1\},B)$, that is, $x\to x$ is a justification of $a\to b\righttherefore c\to d$ iff $a=b$ and $c=d$. Inequality, on the other hand, cannot be detected without negation which is not available in $(\{0,1\})$ (but see \prettyref{§:Negation}). Interestingly, negation is detected \textit{indirectly}.

We have the following result.

\begin{thm}\label{t:B} We have the following table of boolean proportions:
\begin{align*}
\def\thmref#1{\llap{\hyperref[#1]{Thm}}\hyperref[#1]{.\ \ref*{#1}}}
\def\arraystretch{1.1}
\begin{array}{@{}c@{\,\,}c@{\,\,}c@{\,\,}c||c@{\hspace*{10mm}}c||c@{\hspace*{10mm}}c||c@{\hspace*{10mm}}c||c@{\hspace*{13mm}}c@{}}
	a & b & c & d
			& (\{0,1\mathrlap{\})} & \text{Proof} 
	       	& (\{0,1\mathrlap{\},0)} & \text{Proof} 
	       	& (\{0,1\mathrlap{\},1)} & \text{Proof} 
	       	& (\{0,1\mathrlap{\},0,1)} & \text{Proof}\\
	\hline
  \enforceminheight{11pt}
	0 & 0 & 0 & 0 
	       & \mathbf T & \text{\thmref{t:p}}
	       & \mathbf T & \text{\thmref{t:p}} 
	       & \mathbf T & \text{\thmref{t:p}} 
	       & \mathbf T & \text{\thmref{t:p}}\\ 
	1 & 0 & 0 & 0 
	       & \mathbf F & 
	       & \mathbf F & \text{\prettyref{equ:B0_notmodels_0100_1000}}
	       & \mathbf F & \text{\prettyref{equ:B1_notmodels_0100_1000}}
	       & \mathbf F & \text{\prettyref{equ:B01_notmodels_0100_1000}} \\
	0 & 1 & 0 & 0 
	       & \mathbf F & 
	       & \mathbf F & \text{\prettyref{equ:B0_notmodels_0100_1000}}
	       & \mathbf F & \text{\prettyref{equ:B1_notmodels_0100_1000}}
	       & \mathbf F & \text{\prettyref{equ:B01_notmodels_0100_1000}} \\
	1 & 1 & 0 & 0 
	       & \mathbf T & \text{\thmref{t:p}} 
	       & \mathbf T & \text{\thmref{t:p}} 
	       & \mathbf T & \text{\thmref{t:p}} 
	       & \mathbf T & \text{\thmref{t:p}}\\ 
	0 & 0 & 1 & 0 
	       & \mathbf F & 
	       & \mathbf F & \text{\prettyref{equ:B0_notmodels_0001_0010}}
	       & \mathbf F & \text{\prettyref{equ:B1_notmodels_0010_0001}}
	       & \mathbf F & \text{\prettyref{equ:B01_notmodels_0001_0010}} \\
	1 & 0 & 1 & 0 
	       & \mathbf T & \text{\thmref{t:p}} 
	       & \mathbf T & \text{\thmref{t:p}} 
	       & \mathbf T & \text{\thmref{t:p}} 
	       & \mathbf T & \text{\thmref{t:p}}\\ 
	\hline
  \enforceminheight{11pt}
	0 & 1 & 1 & 0 
	       & \mathbf T & 
	       & \mathbf F & \text{\prettyref{equ:B0_notmodels_1001_0110}}
	       & \mathbf F & \text{\prettyref{equ:B1_notmodels_0110_1001}}
	       & \mathbf F & \text{\prettyref{equ:B01_notmodels_0110_1001}} \\
	\hline
  \enforceminheight{11pt}
	1 & 1 & 1 & 0 
	       & \mathbf F & 
	       & \mathbf F & \text{\prettyref{equ:B0_notmodels_1110_1101}}
	       & \mathbf F & \text{\prettyref{equ:B1_notmodels_1110_1101}}
	       & \mathbf F & \text{\prettyref{equ:B01_notmodels_1110_1101}}\\
	0 & 0 & 0 & 1 
	       & \mathbf F & 
	       & \mathbf F & \text{\prettyref{equ:B0_notmodels_0001_0010}}
	       & \mathbf F & \text{\prettyref{equ:B1_notmodels_0010_0001}}
	       & \mathbf F & \text{\prettyref{equ:B01_notmodels_0001_0010}}\\
	\hline
  \enforceminheight{11pt}
	1 & 0 & 0 & 1 
	       & \mathbf T & 
	       & \mathbf F & \text{\prettyref{equ:B0_notmodels_1001_0110}}
	       & \mathbf F & \text{\prettyref{equ:B1_notmodels_0110_1001}}
	       & \mathbf F & \text{\prettyref{equ:B01_notmodels_0110_1001}}\\
	\hline
  \enforceminheight{11pt}
	0 & 1 & 0 & 1 
	       & \mathbf T & \text{\thmref{t:p}} 
	       & \mathbf T & \text{\thmref{t:p}} 
	       & \mathbf T & \text{\thmref{t:p}} 
	       & \mathbf T & \text{\thmref{t:p}}\\ 
	1 & 1 & 0 & 1 
	       & \mathbf F & 
	       & \mathbf F & \text{\prettyref{equ:B0_notmodels_1110_1101}}
	       & \mathbf F & \text{\prettyref{equ:B1_notmodels_1110_1101}}
	       & \mathbf F & \text{\prettyref{equ:B01_notmodels_1110_1101}}\\
	0 & 0 & 1 & 1 
	       & \mathbf T & \text{\thmref{t:p}} 
	       & \mathbf T & \text{\thmref{t:p}} 
	       & \mathbf T & \text{\thmref{t:p}} 
	       & \mathbf T & \text{\thmref{t:p}}\\ 
	1 & 0 & 1 & 1 
	       & \mathbf F & 
	       & \mathbf F & \text{\prettyref{equ:B0_notmodels_1011_0111}}
	       & \mathbf F & \text{\prettyref{equ:B1_notmodels_1011_0111}} 
	       & \mathbf F & \text{\prettyref{equ:B01_notmodels_1011_0111}}\\
	0 & 1 & 1 & 1 
	       & \mathbf F & 
	       & \mathbf F & \text{\prettyref{equ:B0_notmodels_1011_0111}}
	       & \mathbf F & \text{\prettyref{equ:B1_notmodels_1011_0111}}
	       & \mathbf F & \text{\prettyref{equ:B01_notmodels_1011_0111}}\\
	1 & 1 & 1 & 1 
	       & \mathbf T & \text{\thmref{t:p}} 
	       & \mathbf T & \text{\thmref{t:p}} 
	       & \mathbf T & \text{\thmref{t:p}} 
	       & \mathbf T & \text{\thmref{t:p}} 
\end{array}
\end{align*} The above table justifies the following relation:
\begin{align*} 
	(\{0,1\})\models a:b::c:d \quad\Leftrightarrow\quad (a=b \quad\text{and}\quad c=d) \quad\text{or}\quad (a\neq b \quad\text{and}\quad c\neq d).
\end{align*} This implies that in addition to the properties of \prettyref{t:p}, $(\{0,1\})$ satisfies all the properties in \prettyref{§:P} except for monotonicity. The same applies to $(\{0,1\},B)$, for all $B\subseteq\{0,1\}$.
\end{thm}
\begin{proof}[Proof of \prettyref{t:B}] The $(\{0,1\})$-column is an instance of \cite[Theorem 33]{Antic22}.

We proceed with the proof of the $(\{0,1\},0)$-column. Our first observation is:
\begin{align*} 
	\uparrow_{(\{0,1\},0)} 0=\{0,x\} \quad\text{and}\quad \uparrow_{(\{0,1\},0)} 1=\{x\}.
\end{align*} By \prettyref{r:abcd}, this means that all (non-trivial) justifications in $(\{0,1\},0)$ have one of the following forms (recall that $0\to x$ is not a valid justification since $x$ does not occur in $0$):
\begin{align*} 
	0\to 0 \quad\text{or}\quad x\to 0 \quad\text{or}\quad x\to x.
\end{align*} We therefore have:
\begin{align*}
\begin{array}{cccc||l}
	a & b & c & d & \uparrow_{(\{0,1\},0)}(a\to b\righttherefore c\to d)\\[1pt]
	\hline
  \enforceminheight{11pt}
	0 & 0 & 0 & 0 & \{0\to 0,x\to 0,x\to x\}\\
	1 & 0 & 0 & 0 & \{x\to 0\}\\
	0 & 1 & 0 & 0 & \emptyset\\ 
	1 & 1 & 0 & 0 & \{x\to x\}\\
	0 & 0 & 1 & 0 & \{x\to 0\}\\
	1 & 0 & 1 & 0 & \{x\to 0\}\\
	0 & 1 & 1 & 0 & \emptyset\\
	1 & 1 & 1 & 0 & \emptyset\\
	0 & 0 & 0 & 1 & \emptyset\\ 
	1 & 0 & 0 & 1 & \emptyset\\
	0 & 1 & 0 & 1 & \emptyset\\ 
	1 & 1 & 0 & 1 & \emptyset\\
	0 & 0 & 1 & 1 & \{x\to x\}\\
	1 & 0 & 1 & 1 & \emptyset\\
	0 & 1 & 1 & 1 & \emptyset\\
	1 & 1 & 1 & 1 & \{x\to x\}
\end{array}
\end{align*} This implies
\begin{align*}
	&\uparrow_{(\{0,1\},0)}(1\to 1\righttherefore 1\to 0)\subsetneq\ \uparrow_{(\{0,1\},0)}(1\to 1\righttherefore 1\to 1),\\
	&\uparrow_{(\{0,1\},0)}(0\to 0\righttherefore 0\to 1)\subsetneq\ \uparrow_{(\{0,1\},0)}(0\to 0\righttherefore 0\to 0),\\
	&\uparrow_{(\{0,1\},0)}(1\to 0\righttherefore 0\to 1)\subsetneq\ \uparrow_{(\{0,1\},0)}(1\to 0\righttherefore 0\to 0).
\end{align*} This further implies:
\begin{align} 
	\label{equ:B0_notmodels_0100_1000} &(\{0,1\},0)\not\models 0:1::0:0 \quad\text{and}\quad (\{0,1\},0)\not\models 1:0::0:0,\\
	\label{equ:B0_notmodels_0001_0010} &(\{0,1\},0)\not\models 0:0::0:1 \quad\text{and}\quad (\{0,1\},0)\not\models 0:0::1:0,\\
	\label{equ:B0_notmodels_1001_0110} &(\{0,1\},0)\not\models 1:0::0:1 \quad\text{and}\quad (\{0,1\},0)\not\models 0:1::1:0,\\
	\label{equ:B0_notmodels_1110_1101} &(\{0,1\},0)\not\models 1:1::1:0 \quad\text{and}\quad (\{0,1\},0)\not\models 1:1::0:1,\\ 
	\label{equ:B0_notmodels_1011_0111} &(\{0,1\},0)\not\models 1:0::1:1 \quad\text{and}\quad (\{0,1\},0)\not\models 0:1::1:1.
\end{align}\\

We proceed with the proof of the $(\{0,1\},1)$-column, which is analogous to the proof of $(\{0,1\},0)$-column. Our first observation is:
\begin{align*} 
	\uparrow_{(\{0,1\},1)} 0=\{x\} \quad\text{and}\quad \uparrow_{(\{0,1\},1)} 1=\{1,x\}.
\end{align*} By \prettyref{r:abcd}, this means that all non-trivial justifications in $(\{0,1\},1)$ have one of the following forms:
\begin{align*} 
	1\to 1 \quad\text{or}\quad x\to 1 \quad\text{or}\quad x\to x.
\end{align*} We therefore have (this table is analogous to the previous table):
\begin{align*}
\begin{array}{cccc||l}
	a & b & c & d & \uparrow_{(\{0,1\},1)}(a\to b\righttherefore c\to d)\\[1pt]
	\hline
  \enforceminheight{11pt}
	0 & 0 & 0 & 0 & \{x\to x\}\\
	1 & 0 & 0 & 0 & \emptyset\\
	0 & 1 & 0 & 0 & \emptyset\\
	1 & 1 & 0 & 0 & \{x\to x\}\\
	0 & 0 & 1 & 0 & \emptyset\\
	1 & 0 & 1 & 0 & \emptyset\\ 
	0 & 1 & 1 & 0 & \emptyset\\
	1 & 1 & 1 & 0 & \emptyset\\ 
	0 & 0 & 0 & 1 & \emptyset\\
	1 & 0 & 0 & 1 & \emptyset\\
	0 & 1 & 0 & 1 & \{x\to 1\}\\
	1 & 1 & 0 & 1 & \{x\to 1\}\\
	0 & 0 & 1 & 1 & \{x\to x\}\\
	1 & 0 & 1 & 1 & \emptyset\\ 
	0 & 1 & 1 & 1 & \{x\to 1\}\\
	1 & 1 & 1 & 1 & \{1\to 1,x\to 1,x\to x\}
\end{array}
\end{align*} This implies
\begin{align*} 
	&\uparrow_{(\{0,1\},1)}(0\to 1\righttherefore 0\to 0)\subsetneq\ \uparrow_{(\{0,1\},1)}(0\to 1\righttherefore 0\to 1),\\
	&\uparrow_{(\{0,1\},1)}(0\to 1\righttherefore 1\to 0)\subsetneq\ \uparrow_{(\{0,1\},1)}(0\to 1\righttherefore 1\to 1),\\
	&\uparrow_{(\{0,1\},1)}(1\to 1\righttherefore 1\to 0)\subsetneq\ \uparrow_{(\{0,1\},1)}(1\to 1\righttherefore 1\to 1).
\end{align*} This further implies:
\begin{align} 
	\label{equ:B1_notmodels_0100_1000} &(\{0,1\},1)\not\models 0:1::0:0 \quad\text{and}\quad (\{0,1\},1)\not\models 1:0::0:0,\\
	\label{equ:B1_notmodels_0010_0001} &(\{0,1\},1)\not\models 0:0::1:0 \quad\text{and}\quad (\{0,1\},1)\not\models 0:0::0:1,\\ 
	\label{equ:B1_notmodels_0110_1001} &(\{0,1\},1)\not\models 0:1::1:0 \quad\text{and}\quad (\{0,1\},1)\not\models 1:0::0:1,\\ 
	\label{equ:B1_notmodels_1110_1101} &(\{0,1\},1)\not\models 1:1::1:0 \quad\text{and}\quad (\{0,1\},1)\not\models 1:1::0:1,\\
	\label{equ:B1_notmodels_1011_0111} &(\{0,1\},1)\not\models 1:0::1:1 \quad\text{and}\quad (\{0,1\},1)\not\models 0:1::1:1.
\end{align}\\

We proceed with the proof of the $(\{0,1\},0,1)$-column. Our first observation is:
\begin{align*} 
	\uparrow_{(\{0,1\},0,1)} 0=\{0,x\} \quad\text{and}\quad \uparrow_{(\{0,1\},0,1)} 1=\{1,x\}.
\end{align*} By \prettyref{r:abcd}, this means that all non-trivial justifications in $(\{0,1\},0,1)$ have one of the following forms:
\begin{align*} 
	a\to b \quad\text{or}\quad x\to b \quad\text{or}\quad x\to x,\quad\text{for all $a,b\in\{0,1\}$.}
\end{align*} We therefore have (the following table is obtained by joining the previous two and --- since we now have \textit{both} constants 0 and 1 available --- by adding the justifications $0\to 1$ and $1\to 0$.)
\begin{align*}
\begin{array}{cccc||l}
	a & b & c & d & \uparrow_{(\{0,1\},0,1)}(a\to b\righttherefore c\to d)\\[1pt]
	\hline
  \enforceminheight{11pt}
	0 & 0 & 0 & 0 & \{0\to 0,x\to 0,x\to x\}\\
	1 & 0 & 0 & 0 & \{x\to 0\}\\
	0 & 1 & 0 & 0 & \emptyset\\ 
	1 & 1 & 0 & 0 & \{x\to x\}\\
	0 & 0 & 1 & 0 & \{x\to 0\}\\
	1 & 0 & 1 & 0 & \{1\to 0,x\to 0\}\\
	0 & 1 & 1 & 0 & \emptyset\\
	1 & 1 & 1 & 0 & \emptyset\\ 
	0 & 0 & 0 & 1 & \emptyset\\ 
	1 & 0 & 0 & 1 & \emptyset\\
	0 & 1 & 0 & 1 & \{0\to 1,x\to 1\}\\
	1 & 1 & 0 & 1 & \{x\to 1\}\\
	0 & 0 & 1 & 1 & \{x\to x\}\\
	1 & 0 & 1 & 1 & \emptyset\\ 
	0 & 1 & 1 & 1 & \{x\to 1\}\\
	1 & 1 & 1 & 1 &\{1\to 1,x\to 1,x\to x\}
\end{array}
\end{align*} This implies
\begin{align*} 
	&\uparrow_{(\{0,1\},0,1)}(0\to 1\righttherefore 0\to 0)\subsetneq\ \uparrow_{(\{0,1\},0,1)}(0\to 1\righttherefore 0\to 1),\\
	&\uparrow_{(\{0,1\},0,1)}(0\to 1\righttherefore 1\to 0)\subsetneq\ \uparrow_{(\{0,1\},0,1)}(0\to 1\righttherefore 1\to 1),\\
	&\uparrow_{(\{0,1\},0,1)}(1\to 1\righttherefore 1\to 0)\subsetneq\ \uparrow_{(\{0,1\},0,1)}(1\to 1\righttherefore 1\to 1).
\end{align*} This further implies:
\begin{align} 
	\label{equ:B01_notmodels_0100_1000} &(\{0,1\},0,1)\not\models 0:1::0:0 \quad\text{and}\quad (\{0,1\},0,1)\not\models 1:0::0:0,\\
	\label{equ:B01_notmodels_0110_1001} &(\{0,1\},0,1)\not\models 0:1::1:0 \quad\text{and}\quad (\{0,1\},0,1)\not\models 1:0::0:1,\\
	\label{equ:B01_notmodels_1110_1101} &(\{0,1\},0,1)\not\models 1:1::1:0 \quad\text{and}\quad (\{0,1\},0,1)\not\models 1:1::0:1,\\
	\label{equ:B01_notmodels_0001_0010} &(\{0,1\},0,1)\not\models 0:0::0:1 \quad\text{and}\quad (\{0,1\},0,1)\not\models 0:0::1:0,\\
	\label{equ:B01_notmodels_1011_0111} &(\{0,1\},0,1)\not\models 1:0::1:1 \quad\text{and}\quad (\{0,1\},0,1)\not\models 0:1::1:1.
\end{align}
\end{proof}

\section{Negation}\label{§:Negation}

This section studies boolean proportions in the important case where only the unary negation operation and constants are available. Recall from the previous section that $x\to x$ is a justification of $a\to b\righttherefore c\to d$ iff $a=b$ and $c=d$, which shows that $x\to x$ (or, equivalently, $\neg x\to\neg x$ in case negation is available) can detect equality. However, we have also seen that without negation, inequality cannot be analogously explicitly detected (however, we could detect inequality \textit{implicitly} via symmetries). The situation changes in case negation is available, as $x\to\neg x$ (or, equivalently, $\neg x\to x$) is a justification of $a\to b\righttherefore c\to d$ iff $a\neq b$ and $c\neq d$. Hence, given that negation is part of the structure, we can explicitly detect equality and inequality, which is the essence of the following result.

\begin{thm}\label{t:neg} We have the following table of boolean proportions, for every non-empty subset $B$ of $\{0,1\}$:
\begin{align*}
\def\thmref#1{{Thm}. \ref{#1}}
\def\arraystretch{1.1}
\begin{array}{cccc||c|c||c|c}
	a & b & c & d
		& (\{0,1\},\neg) & \text{Proof} 
	    	& (\{0,1\},\neg,B) & \text{Proof}\\
	\hline
  \enforceminheight{11pt}
	0 & 0 & 0 & 0 
	       & \mathbf T & \text{\thmref{t:p}} 
	       & \mathbf T & \text{\thmref{t:p}}\\ 
	1 & 0 & 0 & 0 
	       & \mathbf F & \text{\prettyref{equ:Bneg_notmodels_1000_0100}}
	       & \mathbf F & \text{\prettyref{equ:BnegB_notmodels_0100_1000}}\\
	0 & 1 & 0 & 0 
	       & \mathbf F & \text{\prettyref{equ:Bneg_notmodels_1000_0100}}
	       & \mathbf F & \text{\prettyref{equ:BnegB_notmodels_0100_1000}}\\
	1 & 1 & 0 & 0 
	       & \mathbf T & \text{\thmref{t:p}} 
	       & \mathbf T & \text{\thmref{t:p}}\\ 
	0 & 0 & 1 & 0 
	       & \mathbf F & \text{\prettyref{equ:Bneg_notmodels_0010_0001}}
	       & \mathbf F & \text{\prettyref{equ:BnegB_notmodels_0001_0010}}\\
	1 & 0 & 1 & 0 
	       & \mathbf T & \text{\thmref{t:p}} 
	       & \mathbf T & \text{\thmref{t:p}}\\ 
	0 & 1 & 1 & 0 
	       & \mathbf T & \text{\prettyref{equ:BnegB_models_0110_1001}}
	       & \mathbf T & \text{\prettyref{equ:BnegB_models_0110_1001}}\\
	1 & 1 & 1 & 0 
	       & \mathbf F & \text{\prettyref{equ:Bneg_notmodels_1110_1101}}
	       & \mathbf F & \text{\prettyref{equ:BnegB_notmodels_1110_1101}}\\
	0 & 0 & 0 & 1 
	       & \mathbf F & \text{\prettyref{equ:Bneg_notmodels_0010_0001}}
	       & \mathbf F & \text{\prettyref{equ:BnegB_notmodels_0001_0010}}\\
	1 & 0 & 0 & 1 
	       & \mathbf T & \text{\prettyref{equ:BnegB_models_0110_1001}}
	       & \mathbf T & \text{\prettyref{equ:BnegB_models_0110_1001}}\\
	0 & 1 & 0 & 1 
	       & \mathbf T & \text{\thmref{t:p}} 
	       & \mathbf T & \text{\thmref{t:p}}\\ 
	1 & 1 & 0 & 1 
	       & \mathbf F & \text{\prettyref{equ:Bneg_notmodels_1110_1101}}
	       & \mathbf F & \text{\prettyref{equ:BnegB_notmodels_1110_1101}}\\
	0 & 0 & 1 & 1 
	       & \mathbf T & \text{\thmref{t:p}} 
	       & \mathbf T & \text{\thmref{t:p}}\\ 
	1 & 0 & 1 & 1 
	       & \mathbf F & \text{\prettyref{equ:Bneg_notmodels_1011_0111}}
	       & \mathbf F & \text{\prettyref{equ:BnegB_notmodels_1011_0111}}\\
	0 & 1 & 1 & 1 
	       & \mathbf F & \text{\prettyref{equ:Bneg_notmodels_1011_0111}}
	       & \mathbf F & \text{\prettyref{equ:BnegB_notmodels_1011_0111}}\\
	1 & 1 & 1 & 1 
	       & \mathbf T & \text{\thmref{t:p}} 
	       & \mathbf T & \text{\thmref{t:p}} 
\end{array}
\end{align*} The above table justifies the following relations, for every subset $B$ of $\{0,1\}$:
\begin{align}\label{equ:neg_models_abcd} 
	(\{0,1\},\neg,B)\models a:b::c:d \quad&\Leftrightarrow\quad (a=b \quad\text{and}\quad c=d)\quad\text{or}\quad(a\neq b \quad\text{and}\quad c\neq d)\\
		\quad&\Leftrightarrow\quad (\{0,1\})\models a:b::c:d.
\end{align} This implies that in addition to the properties in \prettyref{t:p}, $(\{0,1\},\neg,B)$ satisfies all the properties in \prettyref{§:P}.\footnote{Monotonicity will be a consequence of the forthcoming \prettyref{t:lor_neg}.}
\end{thm}
\begin{proof}[Proof of \prettyref{t:neg}] We begin with the $(\{0,1\},\neg)$-column. Since $(\{0,1\},\neg)$ contains only the unary function $\neg$ and no constants, our first observation is:
\begin{align*} 
	\uparrow_{(\{0,1\},\neg)} 0=\ \uparrow_{(\{0,1\},\neg)} 1=\{x,\neg x\}.
\end{align*} By \prettyref{r:abcd}, this means that all non-trivial justifications in $(\{0,1\},\neg)$ have one of the following forms:
\begin{align*} 
	x\to x \quad\text{or}\quad x\to\neg x \quad\text{or}\quad \neg x\to x \quad\text{or}\quad \neg x\to\neg x.
\end{align*} We therefore have
\begin{align*} 
	\uparrow_{(\{0,1\},\neg)}(a\to b\righttherefore c\to d)=\begin{cases}
		\{x\to x,\neg x\to\neg x\} & a=b\text{ and }c=d\\
		\{x\to\neg x,\neg x\to x\} & a\neq b\text{ and }c\neq d\\
		\emptyset & \text{otherwise}.
	\end{cases}
\end{align*} This implies
\begin{align*} 
	&\uparrow_{(\{0,1\},\neg)}(1\to 0\righttherefore 0\to 0)\subsetneq\ \uparrow_{(\{0,1\},\neg)}(1\to 0\righttherefore 0\to 1),\\
	&\uparrow_{(\{0,1\},\neg)}(1\to 1\righttherefore 1\to 0)\subsetneq\ \uparrow_{(\{0,1\},\neg)}(1\to 1\righttherefore 1\to 1).
\end{align*} This further implies:
\begin{align} 
	\label{equ:Bneg_notmodels_1000_0100} &(\{0,1\},\neg)\not\models 1:0::0:0 \quad\text{and}\quad (\{0,1\},\neg)\not\models 0:1::0:0,\\
	\label{equ:Bneg_notmodels_0010_0001} &(\{0,1\},\neg)\not\models 0:0::1:0 \quad\text{and}\quad (\{0,1\},\neg)\not\models 0:0::0:1,\\
	\label{equ:Bneg_notmodels_1110_1101} &(\{0,1\},\neg)\not\models 1:1::1:0 \quad\text{and}\quad (\{0,1\},\neg)\not\models 1:1::0:1,\\
	\label{equ:Bneg_notmodels_1011_0111} &(\{0,1\},\neg)\not\models 1:0::1:1 \quad\text{and}\quad (\{0,1\},\neg)\not\models 0:1::1:1.
\end{align}\\

We proceed with the $(\{0,1\},\neg,B)$-column, where $B$ is a non-empty subset of $\{0,1\}$. Notice that as soon as $B$ contains a boolean $a$, its complement $\neg a$ has a `name' in our language --- this immediately implies that the structures $(\{0,1\},\neg,0)$, $(\{0,1\},\neg,1)$, and $(\{0,1\},\neg,0,1)$ entail the same boolean proportions. Without loss of generality, we can therefore assume $B=\{0,1\}$. Our first observation is:
\begin{align*} 
	\uparrow_{(\{0,1\},\neg,B)} 0=\{0,x,\neg x\} \quad\text{and}\quad \uparrow_{(\{0,1\},\neg,B)} 1=\{1,x,\neg x\}.
\end{align*} By \prettyref{r:abcd}, this means that all non-trivial justifications in $(\{0,1\},\neg,B)$ have one of the following forms:
\begin{align*} 
	a\to b,\quad x\to b,\quad \neg x\to b,\quad x\to x,\quad \neg x\to x,\quad x\to \neg x,\quad \neg x\to \neg x,\quad\text{for all $a,b\in B$}.
\end{align*} We therefore have
\begin{align*}
\begin{array}{cccc||l}
	a & b & c & d & \uparrow_{(\{0,1\},\neg,B)}(a\to b\righttherefore c\to d)\\[1pt]
	\hline
  \enforceminheight{11pt}
	0 & 0 & 0 & 0 & \{0\to 0,x\to 0,x\to x,\neg x\to\neg x\}\\
	1 & 0 & 0 & 0 & \{x\to 0,\neg x\to 0\}\\
	0 & 1 & 0 & 0 & \emptyset\\ 
	1 & 1 & 0 & 0 & \{x\to x,\neg x\to\neg x\}\\
	0 & 0 & 1 & 0 & \{x\to 0,\neg x\to 0\}\\
	1 & 0 & 1 & 0 & \{1\to 0,x\to 0,\neg x\to 0,x\to\neg x,\neg x\to x\}\\
	0 & 1 & 1 & 0 & \{x\to\neg x,\neg x\to x\}\\
	1 & 1 & 1 & 0 & \emptyset\\ 
	0 & 0 & 0 & 1 & \emptyset\\ 
	1 & 0 & 0 & 1 & \{x\to\neg x,\neg x\to x\}\\
	0 & 1 & 0 & 1 & \{0\to 1,x\to 1,\neg x\to 1,x\to\neg x,\neg x\to x\}\\
	1 & 1 & 0 & 1 & \{x\to 1,\neg x\to 1\}\\
	0 & 0 & 1 & 1 & \{x\to x,\neg x\to\neg x\}\\
	1 & 0 & 1 & 1 & \emptyset\\ 
	0 & 1 & 1 & 1 & \{x\to 1,\neg x\to 1\}\\
	1 & 1 & 1 & 1 & \{1\to 1,x\to 1,\neg x\to 1,x\to x,\neg x\to\neg x\}
\end{array}
\end{align*} This implies
\begin{align*}
       &\uparrow_{(\{0,1\},\neg,B)}(0\to 1\righttherefore 0\to 0)\subsetneq\ \uparrow_{(\{0,1\},\neg,B)}(0\to 1::0\to 1),\\
       &\uparrow_{(\{0,1\},\neg,B)}(1\to 1\righttherefore 1\to 0)\subsetneq\ \uparrow_{(\{0,1\},\neg,B)}(1\to 1::1\to 1).
\end{align*} This further implies:
\begin{align} 
       \label{equ:BnegB_notmodels_0100_1000} (\{0,1\},\neg,B)\not\models 0:1::0:0 \quad\text{and}\quad (\{0,1\},\neg,B)\not\models 1:0::0:0,\\
       \label{equ:BnegB_notmodels_1110_1101} (\{0,1\},\neg,B)\not\models 1:1::1:0 \quad\text{and}\quad (\{0,1\},\neg,B)\not\models 1:1::0:1,\\
       \label{equ:BnegB_notmodels_0001_0010} (\{0,1\},\neg,B)\not\models 0:0::0:1 \quad\text{and}\quad (\{0,1\},\neg,B)\not\models 0:0::1:0,\\
       \label{equ:BnegB_notmodels_1011_0111} (\{0,1\},\neg,B)\not\models 1:0::1:1 \quad\text{and}\quad (\{0,1\},\neg,B)\not\models 0:1::1:1.
\end{align} As a direct consequence of \prettyref{t:FPT} with $ t(x):=\neg x$, injective in $(\{0,1\},\neg,B)$, we have
\begin{align}\label{equ:BnegB_models_0110_1001} 
	(\{0,1\},\neg,B)\models a:\neg a::c:\neg c,\quad\text{for all $a,c\in\{0,1\}$}.
\end{align}
\end{proof}

\section{Exclusive or}\label{§:XOR}

In this section, we study boolean proportions with respect to the exclusive or operation. Interestingly, it turns out that in case the underlying algebra contains the constant 1, the equation $x+1=\neg x$ shows that we can represent the negation operation, which then yields
\begin{align*} 
	(\{0,1\},+,1)\models a:b::c:d \quad\Leftrightarrow\quad (\{0,1\},\neg)\models a:b::c:d.
\end{align*} However, in case $1$ is not included, the proportions $1:0::1:0$ and $0:1::0:1$ do not hold (see discussion in \prettyref{§:Comparison}).

\begin{thm}\label{t:+} We have the following table of boolean proportions, for all $\{1\}\subseteq B\subseteq\{0,1\}$: 
\begin{align*}
\begin{array}{cccc||cr||c}
	a & b & c & d & (\{0,1\},+) & \text{Proof} &(\{0,1\},+,B)\\
	\hline
  \enforceminheight{11pt} 
	0 & 0 & 0 & 0 & \mathbf T & \text{reflexivity} & \mathbf T\\
	1 & 0 & 0 & 0 & \mathbf F & \text{\prettyref{equ:y_+_not_models_1000}} & \mathbf F\\
	0 & 1 & 0 & 0 & \mathbf F & \text{\prettyref{equ:y_+_not_models_1000}} & \mathbf F\\
	1 & 1 & 0 & 0 & \mathbf T & \text{inner reflexivity} & \mathbf T\\
	0 & 0 & 1 & 0 & \mathbf F & \text{\prettyref{equ:y_+_not_models_1000}} & \mathbf F\\
	1 & 0 & 1 & 0 & \mathbf T & \text{reflexivity} & \mathbf T\\
	\hline
  \enforceminheight{11pt}
	0 & 1 & 1 & 0 & \mathbf F & \text{\prettyref{equ:y_+_not_models_1001}} & \mathbf T\\
	\hline
  \enforceminheight{11pt}
	1 & 1 & 1 & 0 & \mathbf F & \text{\prettyref{equ:y_+_not_models_1110}} & \mathbf F\\
	0 & 0 & 0 & 1 & \mathbf F & \text{\prettyref{equ:y_+_not_models_1000}} & \mathbf F\\
	\hline
  \enforceminheight{11pt}
	1 & 0 & 0 & 1 & \mathbf F & \text{\prettyref{equ:y_+_not_models_1001}} & \mathbf T\\
	\hline
  \enforceminheight{11pt}
	0 & 1 & 0 & 1 & \mathbf T & \text{reflexivity} & \mathbf T\\
	1 & 1 & 0 & 1 & \mathbf F & \text{\prettyref{equ:y_+_not_models_1110}} & \mathbf F\\
	0 & 0 & 1 & 1 & \mathbf T & \text{inner reflexivity} & \mathbf T\\
	1 & 0 & 1 & 1 & \mathbf F & \text{\prettyref{equ:y_+_not_models_1110}} & \mathbf F\\
	0 & 1 & 1 & 1 & \mathbf F & \text{\prettyref{equ:y_+_not_models_1110}} & \mathbf F\\
	1 & 1 & 1 & 1 & \mathbf T & \text{reflexivity} & \mathbf T
\end{array}
\end{align*} The above table justifies the following relations:
\begin{align*} 
	(\{0,1\},+)\models a:b::c:d \quad\Leftrightarrow\quad (\{0,1\},B)\models a:b::c:d,\quad\text{for all $\emptyset\subsetneq B\subseteq\{0,1\}$}
\end{align*} and
\begin{align*} 
	(\{0,1\},+,B)\models a:b::c:d \quad\Leftrightarrow\quad (\{0,1\})\models a:b::c:d,\quad\text{for all $\{1\}\subseteq B\subseteq\{0,1\}$}
\end{align*}
\end{thm}
\begin{proof} Given that we have a single binary function symbol $+$ in our language, terms have the form $a_1x_1+\ldots+a_nx_n$, for some $a_1,\ldots,a_n\in\{0,1\}$ and $n\geq 1$. Since in $(\{0,1\},+)$ we always have
\begin{align}\label{equ:2x=0_3x=x} 
	2x=0 \quad\text{and}\quad 3x=x,
\end{align} terms can always be simplified to the form $x_1+\ldots+x_n$ not containing coefficients. Justifications in $(\{0,1\},+)$ thus have the form $x_1+\ldots+x_n\to x_{i_1}+\ldots+x_{i_k}$, for some \textit{distinct} $i_1,\ldots,i_k\in [1,n]$, $k\leq n$. Notice that we can always rearrange the variables in a justification to obtain the form $x_1+\ldots+x_n\to x_1+\ldots+x_m$, for some $1\leq m\leq n$. Moreover, observe that from \prettyref{equ:2x=0_3x=x}, we can deduce that the only justifications containing a single variable $x$ are 
\begin{align}\label{equ:x_to_x_x_to_2x_}
	x\to x,\quad x\to 2x,\quad 2x\to x,\quad 2x\to 2x.
\end{align} The first justification $x\to x$ justifies inner reflexive arrow proportions of the form $a\to a \righttherefore c\to c$. The second and third justifications $x\to 2x$ and $2x\to x$ justify all arrow proportions of the form $a\to 0 \righttherefore c\to 0$ and $0\to a \righttherefore 0\to c$, where $a$ and $c$ are arbitrary, respectively. Notice that $x\to 2x$ is in fact a characteristic justification of $a\to 0 \righttherefore c\to 0$ which shows
\begin{align}\label{euq:y_+_models_a->0>c->0} 
	(\{0,1\},+)\models a\to 0 \righttherefore c\to 0,\quad\text{for all $a,c\in\{0,1\}$}.
\end{align} Lastly, the fourth justification $2x\to 2x$ justifies only the arrow proportion $0\to 0 \righttherefore 0\to 0$ and it is therefore a characteristic justification.

In the rest of the proof we always assume that $1\leq m\leq n$ and $n\geq 2$.

We now have the following proofs:
\begin{itemize}
\item We wish to prove
\begin{align}\label{equ:y_+_not_models_1000}
	(\{0,1\},+)\not\models 1:0::0:0.
\end{align} By definition of analogical proportions, it suffices to show
\begin{align}\label{equ:y_+_not_models_0001} 
	(\{0,1\},+)\not\models 0\to 0 \righttherefore 0\to 1.
\end{align} We have
\begin{align*} 
	x\to x\in\ \uparrow_{(\{0,1\},+)}(0\to 0 \righttherefore 0\to 0) \quad\text{but}\quad x\to x\not\in\ \uparrow_{(\{0,1\},+)}(0\to 0 \righttherefore 0\to 1).
\end{align*} Since \textit{every} rule $s\to t$ is a justification of $0\to 0 \righttherefore 0\to 0$ by substituting $0$ for all variables in $s$ and $t$, we have thus shown
\begin{align*} 
	\uparrow_{(\{0,1\},+)}(0\to 0 \righttherefore 0\to 1)\subsetneq\ \uparrow_{(\{0,1\},+)}(0\to 0 \righttherefore 0\to 0),
\end{align*} proving \prettyref{equ:y_+_not_models_0001} and thus \prettyref{equ:y_+_not_models_1000}.

\item Next, we wish to prove
\begin{align}\label{equ:y_+_not_models_1110} 
	(\{0,1\},+)\not\models 1:1::1:0,
\end{align} which, by inner symmetry \prettyref{equ:inner_symmetry}, is equivalent to
\begin{align*} 
	(\{0,1\},+)\not\models 1:1::0:1.
\end{align*} For this, it suffices to show
\begin{align}\label{equ:y_+_not_models_1101}
	(\{0,1\},+)\not\models 1\to 1 \righttherefore 0\to 1.
\end{align} We have
\begin{align}\label{equ:x_to_x_not_in_uparrow_y_+_1100} 
	x\to x\in\ \uparrow_{(\{0,1\},+)}(1\to 1 \righttherefore 0\to 0) \quad\text{but}\quad x\to x\not\in\ \uparrow_{(\{0,1\},+)}(1\to 1 \righttherefore 0\to 1).
\end{align} It is easy to check that all the other justifications from \prettyref{equ:x_to_x_x_to_2x_} containing only a single variable satisfy
\begin{align*} 
	x\to 2x,2x\to x,2x\to 2x\not\in\ \uparrow_{(\{0,1\},+)}(1\to 1 \righttherefore 0\to 0)
\end{align*} and
\begin{align*} 
	x\to 2x,2x\to x,2x\to 2x\not\in\ \uparrow_{(\{0,1\},+)}(1\to 1 \righttherefore 0\to 1).
\end{align*} Our next observation is that \textit{every} rule of the form $x_1+\ldots+x_n\to x_1+\ldots+x_m$, containing at least two variables, is a justification of $0\to 0 \righttherefore 1\to 1$:
\begin{center}
\begin{tikzpicture}[node distance=1cm and 0.5cm]
\node (a)               {$1$};
\node (d1) [right=of a] {$\to$};
\node (b) [right=of d1] {$1$};
\node (d2) [right=of b] {$\righttherefore $};
\node (c) [right=of d2] {$0$};
\node (d3) [right=of c] {$\to$};
\node (d) [right=of d3] {$0$.};
\node (s) [below=of b] {$x_1+\ldots+x_n$};
\node (t) [above=of c] {$x_1+\ldots+x_m$};

\draw (a) to [edge label'={$\mathbf x/(1,0,\ldots,0)$}] (s); 
\draw (c) to [edge label={$\mathbf x/(0,\ldots,0)$}] (s);
\draw (b) to [edge label={$\mathbf x/(1,0,\ldots,0)$}] (t);
\draw (d) to [edge label'={$\mathbf x/(0,\ldots,0)$}] (t);
\end{tikzpicture}
\end{center} Together with \prettyref{equ:x_to_x_not_in_uparrow_y_+_1100}, we have thus shown
\begin{align*} 
	\uparrow_{(\{0,1\},+)}(1\to 1 \righttherefore 0\to 1)\subsetneq\ \uparrow_{(\{0,1\},+)}(1\to 1 \righttherefore 0\to 0),
\end{align*} proving \prettyref{equ:y_+_not_models_1101} and thus \prettyref{equ:y_+_not_models_1110}.

\item Finally, we wish to prove
\begin{align}\label{equ:y_+_not_models_1001} 
	(\{0,1\},+)\not\models 1:0::0:1.
\end{align} By definition of analogical proportions, it suffices to prove 
\begin{align}\label{equ:y_+_not_models_1->0>0->1} 
	(\{0,1\},+)\not\models 1\to 0 \righttherefore 0\to 1.
\end{align} We have
\begin{align*} 
	x\to 2x\in\ \uparrow_{(\{0,1\},+)}(1\to 0 \righttherefore 0\to 0) \quad\text{but}\quad x\to 2x\not\in\ \uparrow_{(\{0,1\},+)}(1\to 0 \righttherefore 0\to 1).
\end{align*} We claim
\begin{align}\label{equ:uparrow_y_+_1->0>0->1_subseteq_uparrow_y_+_1->0>0->0} 
	\uparrow_{(\{0,1\},+)}(1\to 0 \righttherefore 0\to 1)\subseteq\ \uparrow_{(\{0,1\},+)}(1\to 0 \righttherefore 0\to 0).
\end{align} The fact
\begin{align*} 
	x\to x,x\to 2x,2x\to 2x\not\in\ \uparrow_{(\{0,1\},+)}(1\to 0 \righttherefore 0\to 1)
\end{align*} shows that there is no justification of $1\to 0 \righttherefore 0\to 1$ from \prettyref{equ:x_to_x_x_to_2x_} containing a single variable. Every justification of $1\to 0 \righttherefore 0\to 1$ containing at least two variables and having the form $x_1+\ldots+x_n\to x_1+\ldots+x_m$, $m<n$, is a justification of $1\to 0 \righttherefore 0\to 0$ as well:
\begin{center}
\begin{tikzpicture}[node distance=1cm and 0.5cm]
\node (a)               {$1$};
\node (d1) [right=of a] {$\to$};
\node (b) [right=of d1] {$0$};
\node (d2) [right=of b] {$\righttherefore $};
\node (c) [right=of d2] {$0$};
\node (d3) [right=of c] {$\to$};
\node (d) [right=of d3] {$0$.};
\node (s) [below=of b] {$x_1+\ldots+x_n$};
\node (t) [above=of c] {$x_1+\ldots+x_m$};

\draw (a) to [edge label'={$\mathbf x/(0,\ldots,0;1,0,\ldots,0)$}] (s); 
\draw (c) to [edge label={$\mathbf x/(0,\ldots,0)$}] (s);
\draw (b) to [edge label={$\mathbf x/(0,\ldots,0;1,0,\ldots,0)$}] (t);
\draw (d) to [edge label'={$\mathbf x/(0,\ldots,0)$}] (t);
\end{tikzpicture}
\end{center} where in $(0,\ldots,0;1,0,\ldots,0)$ the single $1$ occurs at position $m+1$. For $m=n$, we obtain the rule $x_1+\ldots+x_n\to x_1+\ldots+x_n$ justifying only inner reflexive arrow proportions of the form $a\to a \righttherefore c\to c$ irrelevant here. Hence, we have shown \prettyref{equ:uparrow_y_+_1->0>0->1_subseteq_uparrow_y_+_1->0>0->0} which implies \prettyref{equ:y_+_not_models_1->0>0->1} from which \prettyref{equ:y_+_not_models_1001} follows.
\end{itemize}

Finally, the identity $\neg x=x+1$ shows that if the algebra contains the constant $1$, the justification $x\to x+1$ is a characteristic justification of $1:0::0:1$ and $0:1::1:0$.
\end{proof}

\section{Propositional Logic}\label{§:PL}

In this section, we study the boolean structures $(\{0,1\},\lor,\neg)$, $(\{0,1\},\lor,\neg,0)$, $(\{0,1\},\lor,\neg,1)$, and $(\{0,1\},\lor,\neg,0,1)$, where we can employ full propositional logic. Surprisingly, it turns out that these structures are equivalent with respect to boolean proportions to the structure $(\{0,1\})$ of \prettyref{§:Constants} containing only the boolean universe, and the structures of \prettyref{§:Negation} containing only negation. This is the essence of the forthcoming \prettyref{t:lor_neg}. This is interesting as it shows that boolean proportions using full propositional logic can be reduced to the algebra $(\{0,1\})$ containing no functions and no constants.  The proof of \prettyref{t:lor_neg} is, however, different from the proof of \prettyref{t:neg} since computing all generalizations of a boolean element in $(\{0,1\},\lor,\neg,B)$, which amounts to computing all satisfiable or falsifiable propositional formulas, is difficult \cite{Derschowtiz03}.

We have the following result:

\begin{thm}\label{t:lor_neg} We have the following table of boolean proportions, for every subset $B$ of $\{0,1\}$:
\begin{align*}
\def\thmref#1{\llap{Thm}. \ref{#1}}
\def\arraystretch{1.1}
\begin{array}{cccc||c@{\hspace*{10mm}}r}
	a & b & c & d & (\{0,1\},\lor,\neg,B) & \text{Proof}\\
	\hline
	0 & 0 & 0 & 0 
	       & \mathbf T & \text{\thmref{t:p}}\\ 
	1 & 0 & 0 & 0 
	       & \mathbf F & \text{\prettyref{equ:lor_neg_1000_0100}}\\
	0 & 1 & 0 & 0 
	       & \mathbf F & \text{\prettyref{equ:lor_neg_1000_0100}}\\
	1 & 1 & 0 & 0 
	       & \mathbf T & \text{\thmref{t:p}}\\ 
	0 & 0 & 1 & 0 
	       & \mathbf F & \text{\prettyref{equ:lor_neg_0001_0010}}\\
	1 & 0 & 1 & 0 
	       & \mathbf T & \text{\thmref{t:p}}\\ 
	0 & 1 & 1 & 0 
	       & \mathbf T & \text{\prettyref{equ:BlornegB_models_0110_1001}}\\
	1 & 1 & 1 & 0 
	       & \mathbf F & \text{\prettyref{equ:lor_neg_1110_1101}}\\
	0 & 0 & 0 & 1 
	       & \mathbf F & \text{\prettyref{equ:lor_neg_0001_0010}}\\
	1 & 0 & 0 & 1 
	       & \mathbf T & \text{\prettyref{equ:BlornegB_models_0110_1001}}\\
	0 & 1 & 0 & 1 
	       & \mathbf T & \text{\thmref{t:p}}\\ 
	1 & 1 & 0 & 1 
	       & \mathbf F & \text{\prettyref{equ:lor_neg_1110_1101}}\\
	0 & 0 & 1 & 1 
	       & \mathbf T & \text{\thmref{t:p}}\\ 
	1 & 0 & 1 & 1 
	       & \mathbf F & \text{\prettyref{equ:lor_neg_0111_1011}}\\
	0 & 1 & 1 & 1 
	       & \mathbf F & \text{\prettyref{equ:lor_neg_0111_1011}}\\
	1 & 1 & 1 & 1 
	       & \mathbf T & \text{\thmref{t:p}} 
\end{array}
\end{align*} Hence, we have
\begin{align*}
	(\{0,1\},\lor,\neg,B)\models a:b::c:d \quad&\Leftrightarrow\quad (a=b \quad\text{and}\quad c=d)\quad\text{or}\quad(a\neq b \quad\text{and}\quad c\neq d)\\ 
		\quad&\Leftrightarrow\quad (\{0,1\},\neg,B)\models a:b::c:d\\
		\quad&\Leftrightarrow\quad (\{0,1\})\models a:b::c:d.
\end{align*} This implies that in addition to the properties in \prettyref{t:p}, $(\{0,1\},\lor,\neg,B)$ satisfies all the proportional properties in \prettyref{§:P}.
\end{thm}
\begin{proof} We have the following proofs:
\begin{itemize}
	\item According to \prettyref{t:FPT}, the justification $x\to\neg x$ is a characteristic justification of
	\begin{align}\label{equ:BlornegB_models_0110_1001} 
		(\{0,1\},\lor,\neg,B)\models 0:1::1:0 \quad\text{and}\quad (\{0,1\},\lor,\neg,B)\models 1:0::0:1,
	\end{align} since $\neg x$ satisfies the injectivity property presupposed in \prettyref{t:FPT}.

	\item We wish to prove
	\begin{align}\label{equ:lor_neg_1000_0100} 
    \begin{array}{c}
		(\{0,1\},\lor,\neg,B)\not\models 1:0::0:0 \quad\text{and}
    \\
    \quad (\{0,1\},\lor,\neg,B)\not\models 0:1::0:0,\quad\text{for any $B\subseteq\{0,1\}$}.
    \end{array}
	\end{align} If $s\xrightarrow{\mathbf o\to\mathbf u} t$ is a justification of $0\to 1\righttherefore 0\to 0$ in $(\{0,1\},\lor,\neg,B)$, then $s\xrightarrow{\mathbf o\to\mathbf o} t$ is a justification of $0\to 1\righttherefore 0\to 1$ in $(\{0,1\},\lor,\neg,B)$. This shows
	\begin{align*} 
		\uparrow_{(\{0,1\},\lor,\neg,B)}(0\to 1\righttherefore 1\to 0)\subseteq\ \uparrow_{(\{0,1\},\lor,\neg,B)}(0\to 1\righttherefore 0\to 1).
	\end{align*} On the other hand, we have
	\begin{align*} 
		x\to\neg x\in\ \uparrow_{(\{0,1\},\lor,\neg,B)}(0\to 1\righttherefore 0\to 1)
	\end{align*} whereas
	\begin{align*} 
		x\to\neg x\not\in\ \uparrow_{(\{0,1\},\lor,\neg,B)}(0\to 1\righttherefore 0\to 0).
	\end{align*} This shows
	\begin{align*} 
		\uparrow_{(\{0,1\},\lor,\neg,B)}(0\to 1\righttherefore 0\to 0)\subsetneq\ \uparrow_{(\{0,1\},\lor,\neg,B)}(0\to 1\righttherefore 0\to 1).
	\end{align*} from which \prettyref{equ:lor_neg_1000_0100} follows. An analogous argument proves:
	\begin{align}\label{equ:lor_neg_0111_1011} 
    \begin{array}{c}
		(\{0,1\},\lor,\neg,B)\not\models 0:1::1:1 \quad\text{and}\quad
    \\
    (\{0,1\},\lor,\neg,B)\not\models 1:0::1:1,\quad\text{for any $B\subseteq\{0,1\}$}. 
    \end{array}
	\end{align}

	\item We now wish to prove
	\begin{align}\label{equ:lor_neg_1110_1101} 
    \begin{array}{c}
		(\{0,1\},\lor,\neg,B)\not\models 1:1::1:0 \quad\text{and}\quad 
    \\
    (\{0,1\},\lor,\neg,B)\not\models 1:1::0:1,\quad\text{for any $B\subseteq\{0,1\}$}. 
    \end{array}
	\end{align} If $s\xrightarrow{\mathbf o\to\mathbf u} t$ is a justification of $1\to 1\righttherefore 1\to 0$ in $(\{0,1\},\lor,\neg,B)$, then $s\xrightarrow{\mathbf o\to\mathbf o} t$ is a justification of $1\to 1\righttherefore 1\to 1$ in $(\{0,1\},\lor,\neg,B)$. This shows
	\begin{align*} 
		\uparrow_{(\{0,1\},\lor,\neg,B)}(1\to 1\righttherefore 1\to 0)\subseteq\ \uparrow_{(\{0,1\},\lor,\neg,B)}(1\to 1\righttherefore 1\to 1).
	\end{align*} On the other hand, we have
	\begin{align*} 
		x\to x\in\ \uparrow_{(\{0,1\},\lor,\neg,B)}(1\to 1\righttherefore 1\to 1)
	\end{align*} whereas
	\begin{align*} 
		x\to x\not\in\ \uparrow_{(\{0,1\},\lor,\neg,B)}(1\to 1\righttherefore 1\to 0).
	\end{align*} This shows
	\begin{align*} 
		\uparrow_{(\{0,1\},\lor,\neg,B)}(1\to 1\righttherefore 1\to 0)\subsetneq\ \uparrow_{(\{0,1\},\lor,\neg,B)}(1\to 1\righttherefore 1\to 1).
	\end{align*} from which \prettyref{equ:lor_neg_1110_1101} follows. An analogous argument proves:
	\begin{align}\label{equ:lor_neg_0001_0010} 
		(\{0,1\},\lor,\neg,B)\not\models 0:0::0:1 \quad\text{and}\quad (\{0,1\},\lor,\neg,B)\not\models 0:0::1:0.
	\end{align}
\end{itemize}

\end{proof}

\section{Comparison}\label{§:Comparison}

Boolean proportions have been embedded into a logical setting before by \cite{Klein82} and \cite{Miclet09} (and see \cite{Lepage03}). The most important conceptual difference between our framework and the models in \cite{Klein82} and \cite{Miclet09} is that in ours, we make the underlying structure explicit, which allows us to finer distinguish between boolean structures with different functions and constants. Moreover, in contrast to the aforementioned works which define analogical proportions only in the boolean domain, our framework is an instance of an abstract algebraic model formulated in the general language of universal algebra \cite{Antic22}.

Specifically, \cite{Klein82} gives the following definition of boolean proportion:
\begin{align}\label{equ:abcd_Klein82} 
	a:b::_Kc:d \quad:=\quad (a\equiv b)\equiv (c\equiv d).
\end{align} Miclet's \& Prade's \cite{Miclet09}, on the other hand, define boolean proportions as follows: 
\begin{align*} 
	a:b::_{MP}c:d \quad:=\quad ((a\equiv b)\equiv (c\equiv d))\land ((a\text{ xor }b)\supset (a\equiv c)).
\end{align*} 

The following table summarizes the situations in \cite{Miclet09}, \cite{Klein82}, and our framework in $(\{0,1\},+)$ (\prettyref{§:XOR}) and $(\{0,1\},\lor,\neg,B)$ (\prettyref{§:PL}), and it provides arguments for its differences (highlighted lines):
\begin{align*}
\begin{array}{cccc||c|c||c|c}
	a & b & c & d & a:b::_Kc:d & (\{0,1\},\lor,\neg,B) & a:b::_{MP}c:d & (\{0,1\},+)\\
	\hline
  \enforceminheight{11pt}
	0 & 0 & 0 & 0 & \mathbf T & \mathbf T & \mathbf T & \mathbf T \\
	1 & 0 & 0 & 0 & \mathbf F & \mathbf F & \mathbf F & \mathbf F \\
	0 & 1 & 0 & 0 & \mathbf F & \mathbf F & \mathbf F & \mathbf F \\
	1 & 1 & 0 & 0 & \mathbf T & \mathbf T & \mathbf T & \mathbf T \\
	0 & 0 & 1 & 0 & \mathbf F & \mathbf F & \mathbf F & \mathbf F \\
	1 & 0 & 1 & 0 & \mathbf T & \mathbf T & \mathbf T & \mathbf T \\
	\hline
  \enforceminheight{11pt}
	0 & 1 & 1 & 0 & \mathbf T & \mathbf T & \mathbf F & \mathbf F \\
	\hline
  \enforceminheight{11pt}
	1 & 1 & 1 & 0 & \mathbf F & \mathbf F & \mathbf F & \mathbf F \\
	0 & 0 & 0 & 1 & \mathbf F & \mathbf F & \mathbf F & \mathbf F \\
	\hline
  \enforceminheight{11pt}
	1 & 0 & 0 & 1 & \mathbf T & \mathbf T & \mathbf F & \mathbf F \\
	\hline
  \enforceminheight{11pt}
	0 & 1 & 0 & 1 & \mathbf T & \mathbf T & \mathbf T & \mathbf T \\
	1 & 1 & 0 & 1 & \mathbf F & \mathbf F & \mathbf F & \mathbf F \\
	0 & 0 & 1 & 1 & \mathbf T & \mathbf T & \mathbf T & \mathbf T \\
	1 & 0 & 1 & 1 & \mathbf F & \mathbf F & \mathbf F & \mathbf F \\
	0 & 1 & 1 & 1 & \mathbf F & \mathbf F & \mathbf F & \mathbf F \\
	1 & 1 & 1 & 1 & \mathbf T & \mathbf T & \mathbf T & \mathbf T
\end{array}
\end{align*}

Surprisingly, Klein's notion of boolean proportions in \prettyref{equ:abcd_Klein82} coincides with our notion evaluated in $(\{0,1\})$, $(\{0,1\},\neg,B)$, $(\{0,1\},+,1)$, and $(\{0,1\},\lor,\neg,B)$, where $B$ is any set subset of $\{0,1\}$. Recall from \prettyref{t:B} that in $(\{0,1\})$ we have
\begin{align*} 
	(\{0,1\})\models a:b::c:d \quad\Leftrightarrow\quad (a=b \quad\text{and}\quad c=d) \quad\text{or}\quad (a\neq b \quad\text{and}\quad c\neq d).
\end{align*} In Theorems \ref{t:neg} and \ref{t:lor_neg} we have further derived
\begin{align*}
	(\{0,1\},\lor,B)\models a:b::c:d \quad\Leftrightarrow\quad (\{0,1\})\models a:b::c:d,\\
	(\{0,1\},\lor,\neg,B)\models a:b::c:d \quad\Leftrightarrow\quad (\{0,1\})\models a:b::c:d.
\end{align*} This is equivalent to Klein's definition \prettyref{equ:abcd_Klein82}. 

\cite{Miclet09}, on the other hand, do not consider the proportions
\begin{align*} 
	0:1::1:0 \quad\text{and}\quad 1:0::0:1
\end{align*} to be in boolean proportion, `justified' on page 642 as follows:
\begin{quote} 
	The two other cases, namely $0:1::1:0$ and $1:0::0:1$, do not fit the idea that $a$ is to $b$ as $c$ is to $d$, since the changes from $a$ to $b$ and from $c$ to $d$ are not in the same sense. They in fact correspond to cases of maximal analogical dissimilarity, where `$d$ is not at all to $c$ what $b$ is to $a$', but rather `$c$ is to $d$ what $b$ is to $a$'. It emphasizes the non symmetry of the relations between $b$ and $a$, and between $d$ and $c$.
\end{quote} Arguably, this is counter-intuitive as in case negation is available (which it implicitly is in \cite{Miclet09}), given the injectivity (and simplicity) of the negation operation, it is reasonable to conclude that `$a$ is to its negation $\neg a$ what $c$ is to its negation $\neg c$', and $x\to\neg x$ (or, equivalently, $\neg x\to x$) is therefore a plausible \textit{characteristic} justification of $0:1::1:0$ and $1:0::0:1$ in our framework. 

To summarize, our framework differs substantially from the aforementioned models:
\begin{enumerate}
	\item Our model is algebraic in nature and it is naturally embedded within a general model of analogical proportions formulated within the general language of universal algebra.

	\item In our model, we make the underlying universe and its functions and constants explicit, which allows us to make fine distinctions between different boolean domains.

\end{enumerate}

\section{Future Work}\label{§:FW}

Boolean functions are at the core of hardware design and computer science in general \cite{Crama11}. Studying boolean function proportions therefore is a major line of future research directly building on the results of this paper. Since finite sets can be identified with their characteristic boolean functions, studying boolean function proportions is essentially the same as studying finite set proportions. Although \cite[§5.1]{Antic22} provides some first elementary results in that direction (and see \cite[§4.2]{Lepage03}), a complete understanding of set proportions is missing.

More broadly speaking, it is interesting to study analogical proportions in different kinds of mathematical structures as, for example, semigroups and groups, lattices, et cetera --- \cite{Antic22-2} is a first step along these lines studying analogical proportions in monounary algebras. In the context of artificial intelligence it is particularly interesting to study analogical proportions between more sophisticated objects such as, for example, programs, neural networks, automata, et cetera. A recent paper in that direction is the study of logic program proportions for logic program synthesis via analogy-making in \cite{Antic23-23}.

From a mathematical point of view, relating boolean proportions to other concepts of boolean and universal algebra and related subjects is an interesting line of research. For this it will be essential to study the relationship between properties of elements like being ``neutral'' or ``absorbing'' and their proportional properties. At this point --- due to the author's lack of expertise --- it is not entirely clear how boolean and analogical proportions fit into the overall landscape of boolean and universal algebra and relating analogical proportions to other concepts of algebra and logic is therefore an important line of future research. 


\section{Conclusion}

This paper studied boolean proportions by instantiating the abstract algebraic model of analogical proportions in \cite{Antic22} in the boolean setting. It turned out that our model has appealing mathematical properties. Surprisingly, we found that our model coincides with Klein's model \cite{Klein82} in boolean domains containing either negation or no constants, and with Miclet's \& Prade's model \cite{Miclet09} in the remaining boolean domains. That is, our model captures two different approaches to boolean proportions in a single framework. This is particularly interesting as our model is an instance of a general model not explicitly geared towards the boolean setting, which provides further evidence for the applicability of the underlying general framework.

\bibliographystyle{alphaurl}
\bibliography{Bibliography,Publications_J,Submitted}
\end{document}